\documentclass[preprint,12pt,authoryear]{elsarticle}

\usepackage{natbib}
\usepackage{multirow}
\usepackage{amsmath}
\usepackage{amssymb}
\usepackage{graphicx}
\usepackage{hyperref}
\usepackage{subfigure}
\usepackage{soulutf8}

\usepackage[title]{appendix}
\usepackage{lineno}
\usepackage{placeins}
\usepackage{graphicx}
\usepackage{float}
\usepackage{xcolor}
\sethlcolor{yellow}
\usepackage{geometry}

\usepackage{rotating}
\usepackage{subcaption}
\usepackage{pifont}

\soulregister\cite7
\soulregister\ref7
\soulregister\citep7

\journal{TRC}

\begin{document}

\begin{frontmatter}



\title{SemaPop: Semantic-Persona Conditioned and Controllable Population Synthesis} 

\author[label1]{Zhenlin Qin} 
\author[label1]{Yancheng Ling} 
\author[label3]{Leizhen Wang} 
\author[label4]{Francisco Câmara Pereira} 
\author[label1,label2]{Zhenliang Ma\corref{cor1}} 

\cortext[cor1]{Corresponding author.}

\affiliation[label1]{organization={Department of Civil and Architectural Engineering, KTH Royal Institute of Technology},
            country={Sweden}}

\affiliation[label2]{organization={Digital Future, KTH Royal Institute of Technology},
country={Sweden}}

\affiliation[label3]{organization={Department of Data Science and Artificial Intelligence, Monash University},
            country={Australia}}

\affiliation[label4]{organization={Department of Technology, Management and Economics, Technical University of Denmark},
            country={Denmark}}


\begin{abstract}
Population synthesis is essential for individual-level simulation in transport planning and socio-economic analysis, yet remains challenging due to the need to capture both statistical dependencies and high-level behavioral semantics. Existing data-driven approaches predominantly rely on unconditional generation, limiting their ability to support scenario-driven or target-oriented population synthesis. This study proposes SemaPop, a semantic-conditioned and controllable population synthesis framework that introduces persona representations as conditioning signals for generation. By deriving persona text from survey data using large language models (LLMs) and encoding it into semantic embeddings, SemaPop enables controllable population generation under statistical constraints. We instantiate the framework using a GAN-based architecture with marginal regularization to preserve distributional consistency. Extensive experiments demonstrate that SemaPop substantially improves generative performance, yielding closer alignment with target marginal and joint distributions while maintaining sample-level feasibility and diversity under semantic conditioning. Counterfactual analyses further demonstrate that semantic interventions induce systematic and interpretable shifts in generated populations. These results highlight the potential of persona-based semantic conditioning for controllable and scenario-oriented population synthesis.
\end{abstract}


\end{frontmatter}



\section{Introduction}
\label{intro}
Modern socio-economic systems, such as transportation systems and urban systems, are increasingly governed by complex interactions between individual behaviors, demographic structures, and policy interventions \citep{oecd2017systems,helbing2009managing,batty2013new}. Effective planning and policy evaluation in these systems require not only aggregate-level indicators, but also a fine-grained understanding of how heterogeneous individuals respond to changing socio-economic conditions. To capture such heterogeneity and interaction effects, many socio-economic modeling frameworks adopt individual-level simulation paradigms, such as agent-based modeling (ABM) and micro-simulation, which explicitly represent individuals and their behaviors \citep{bonabeau2002agent,gilbert2005simulation,miller2018agent,wang2025agentic}. These approaches critically rely on the availability of synthetic populations, which serve as the foundational layer for initializing individual agents with realistic demographic and behavioral attributes. Consequently, the quality and expressiveness of population synthesis directly determine the credibility of downstream simulations and policy analysis.

Despite their central role in individual-level simulation, population synthesis remains challenging due to the need to integrate heterogeneous types of information. Beyond reproducing marginal distributions of structured variables, realistic synthetic populations must also represent higher-level population semantics that jointly emerge from demographic structures, household contexts, and behavioral tendencies, many of which are only implicitly encoded in survey data. Such latent heterogeneity is difficult to capture within conventional population synthesis frameworks that primarily operate on structured attributes and statistical constraints, motivating the incorporation of semantic abstractions as a complementary layer to statistical population modeling.

Traditional population synthesis methods primarily adopt statistical constraint-driven paradigms. Marginal-based methods, including Iterative Proportional Fitting (IPF), Iterative Proportional Updating (IPU), and copula-based approaches, extrapolate population distributions through empirical marginals or dependency structures \citep{beckman1996creating,norman1999putting,jeong2016copula}. However, these methods remain tied to historical observations and are unable to represent scenario-driven behavioral or demographic shifts. Combinatorial optimization approaches generate synthetic populations by matching predefined constraints, yet do not learn underlying microdata distributions and thus exhibit limited generalization beyond observed conditions \citep{hermes2012review,muller2010population,barthelemy2013synthetic}. 
Recent deep generative models improve statistical realism by learning latent representations from survey and tabular microdata, but remain confined to statistical pattern learning and lack mechanisms to encode semantic context or scenario-specific assumptions \citep{aemmer2022generative,kim2023deep,garrido2020prediction}. This limitation motivates the incorporation of external semantic knowledge, such as persona representations derived from natural language descriptions.

Large Language Models (LLMs) offer a natural mechanism for incorporating external semantic knowledge into population synthesis due to their strong capacity for semantic abstraction \citep{brown2020language,bommasani2021opportunities}. Recent population synthesis research shows that LLMs capture semantic relationships among socio-demographic attributes, helping distinguish plausible, rare, and infeasible combinations \citep{lim2026large}. While survey microdata provide structured statistical attributes, such attributes alone are often insufficient to represent higher-level semantics related to lifestyles, preferences, and behavioral tendencies. By transforming such information into continuous persona embeddings, LLMs provide a compact semantic representation that complements structured tabular variables. Integrating persona embeddings with population synthesis models originally designed without semantic conditioning enables semantic conditioning under statistical constraints, allowing higher-level population semantics to influence generation while preserving distributional consistency.


To better represent high-level population semantics within synthetic population generation, we propose \emph{SemaPop}, a semantic-conditioned population synthesis framework that integrates LLM-derived persona representations with a generative population model under explicit distributional constraints. The framework is instantiated as \emph{SemaPop-GAN}, in which persona embeddings derived from survey records are used to condition the generation process, while marginal regularization is introduced to preserve consistency with target population distributions. Such design enables synthetic populations to remain statistically coherent while allowing semantic variations to influence generated demographic, household, and behavioral characteristics. This study uses synthetic population microdata as a controlled experimental setting for examining semantic conditioning and statistical consistency in population generation. The main contributions of this work are summarized as follows:
\begin{itemize}
\item We introduce a new \emph{semantic-conditioned population synthesis paradigm}, which shifts from conventional data-driven unconditional generation toward controllable population generation via semantic conditioning.

\item We propose \emph{SemaPop}, a semantic-conditioned population synthesis framework that incorporates persona-level representations as conditioning signals while preserving statistical consistency through marginal regularization.

\item We introduce a counterfactual analysis framework at both the semantic embedding and text levels to diagnose the effect of semantic interventions on generated population distributions.

\item We conduct comprehensive empirical evaluations on synthetic population survey data, demonstrating improved statistical realism, semantic controllability, and robustness under post-hoc marginal calibration.
\end{itemize}

The remainder of this paper is structured as follows: Section \ref{related_work} reviews the studies related to population synthesis models, generative modeling with semantic information, and counterfactual analysis. Sections \ref{sec_prel} defines the studied problem and necessary notations. Section \ref{method} proposes the SemaPop framework as well as its instantiation, and then introduce the counterfactual intervention framework. Section \ref{sec_ex} conduct extensive experiments for validating model performance, robustness and effectiveness of persona conditions. The final section concludes the main findings and discusses future work.

\section{Related Work}
\label{related_work}
\subsection{Population Synthesis with Structured Data}
Traditional population synthesis methods primarily rely on constraint-driven or probabilistic formulations operating on structured survey data. Marginal-based approaches, such as IPF and its extensions such as IPU, generate synthetic populations by reweighting or scaling sample records to match externally provided demographic, household, or geographic marginals \citep{birkin1988synthesis,norman1999putting,beckman1996creating,ye2009methodology}. Combinatorial optimization (CO) methods instead formulate synthesis as a discrete selection problem, searching for an optimal combination of individuals that minimizes discrepancies with target constraints through heuristic algorithms such as simulated annealing or genetic optimization \citep{williamson1998estimation,voas2000evaluation,harland2012creating,barthelemy2013synthetic}. Probabilistic and graphical models including Bayesian, mixture-based, and copula-based formulations, explicitly estimate joint distributions by factorizing dependencies or separating marginal and dependence structures, enabling richer modeling of multi-attribute relationships than purely marginal-fitting methods \citep{pritchard2012advances,jeong2016copula,sun2018hierarchical}. Despite differences in formulation, these traditional approaches remain fundamentally constrained by observed data and predefined constraints, limiting their ability to generalize beyond historical distributions or support scenario-driven future population projection.

Data-driven population synthesis has emerged as an alternative paradigm that learns latent distributions and dependency structures directly from individual-level survey data. Early works include Bayesian networks that explicitly model joint demographic and household distributions \citep{sun2015bayesian}. More recently, deep generative models such as variational autoencoders (VAEs) and generative adversarial networks (GANs), have been adapted from image generation to population and transportation applications, enabling flexible modeling of high-dimensional attribute spaces \citep{kingma2013auto,goodfellow2014generative,borysov2019generate,johnsen2022population}. Representative studies apply VAEs to capture latent preference mechanisms \citep{boquet2020variational,borysov2021introducing} and Wasserstein GANs to address challenges such as sampling bias and structural zeros in large-scale population synthesis \citep{garrido2020prediction,kim2023deep}. Hybrid approaches further extend data-driven population synthesis by integrating complementary modeling paradigms. For example, deep generative models have been combined with copula-based dependence structures to improve spatial transferability \citep{jutras2024copula}. Recent work incorporates LLMs with Bayesian networks to capture semantic relationships among socio-demographic attributes and enhance the feasibility of generated populations \citep{lim2026large}.

Compared to traditional marginal-based approaches and CO methods, these data-driven models learn joint population distributions directly from real survey data, thereby improving statistical realism. While they significantly enhance realism and scalability, they predominantly operate as unconditional generation frameworks that aim to approximate the observed population distribution. As a result, the unconditional generation paradigm underlying these methods fundamentally precludes explicit control, limiting their applicability for scenario-driven or target-oriented population synthesis.

\subsection{Semantic Conditioning and Counterfactual Analysis in Generative Models}

Recent advances in generative modeling have increasingly emphasized semantic information as an explicit and controllable factor in the generation process. Across architectures such as VAEs, GANs, and diffusion models, high-level semantic representations are introduced as conditioning signals that guide generation beyond low-level statistical pattern learning \citep{yu2022survey}. These representations are typically integrated into latent spaces or intermediate feature layers, enabling partial disentanglement between semantic intent and underlying statistical structure \citep{harvey2021conditional,rombach2022high,perez2018film}. This line of work highlights a general principle: semantic information can act as an independent signal that systematically shapes generative distributions, improving controllability without requiring explicit modification of the data space. While such semantic conditioning enhances controllability, it also raises an important question: whether and how semantic information actively influences the learned generative distribution. Standard evaluation metrics alone are insufficient to answer this question, as they do not distinguish between semantic effects and underlying statistical regularities. To address this, recent work in machine learning has adopted counterfactual analysis as a tool for probing the role of semantic information in generative models.

Unlike classical counterfactual analysis in causal inference, which relies on explicit structural assumptions \citep{rubin1974estimating,pearl2009causality}, counterfactual analysis in generative modeling is typically operationalized as controlled interventions on inputs or learned representations \citep{karimi2021algorithmic,verma2024counterfactual}. By systematically modifying semantic inputs and examining resulting changes in outputs, these approaches provide an interpretable mechanism for analyzing model behavior and sensitivity.

Two complementary forms of intervention are commonly considered. First, semantic-level interventions operate directly in latent or embedding spaces, where high-level attributes can often be approximated as directions, enabling controlled manipulation of semantic factors \citep{kim2018tcav,shen2020interpreting,harkonen2020ganspace}. Second, text-level interventions modify natural language inputs to induce targeted semantic changes, providing a more interpretable interface for controlling model behavior \citep{madaan2021generate,seifert2024ceval}. These developments are particularly relevant for population synthesis. Behavioral preferences and lifestyle characteristics are inherently semantic, yet are often only weakly captured by structured survey attributes, creating a gap between statistical representation and behavioral interpretation \citep{kim2023deep}. Incorporating persona-based semantic representations helps bridge this gap, while counterfactual interventions provide a principled way to evaluate whether such semantic information actively shapes generated population distributions.

\section{Problem Definition}
\label{sec_prel}

Population synthesis in this study is formulated on individual-level survey data in tabular form, where each row represents a population agent characterized by multi-dimensional attributes, including demographic, household, and behavioral information. Unlike general tabular data generation, population synthesis does not aim to reproduce an identical replica of the observed sample. Instead, it seeks to generate a synthetic population whose individuals remain realistic at the micro level while preserving consistency with population-level statistical patterns, such as marginal distributions and attribute dependencies.

Conventional population synthesis primarily serves the purpose of \emph{population reconstruction}: generating a representative population from limited survey samples. However, in many transport applications, reconstruction alone is insufficient. Analysts often need to generate populations that satisfy \emph{targeted conditions} or \emph{counterfactual assumptions}, for example to represent specific traveler groups, evaluate policy-sensitive subpopulations, test future demographic or behavioral shifts, or perform stress testing under rare but policy-relevant scenarios. These applications require not only statistical realism, but also the ability to \emph{control} the generative process toward desired population profiles. This motivates the need for a new population synthesis framework that is not merely distribution-matching, but also distribution-controllable.

\subsection{Preliminary}

\emph{Population agent}: A population agent $x_i$ is an element of a population set $X$, where $i \in \{1,2,\dots,N\}$. Each agent is represented by a vector of categorical and numerical attributes extracted from individual-level surveys, covering demographic, household, and behavioral characteristics. Formally, each agent $x_i$ is composed of a set of attributes indexed by $j \in \mathcal{J}$, where $\mathcal{J}$ denotes the collection of variables considered in the population representation. These variables can be partitioned into categorical attributes $\mathcal{J}_{\mathrm{cat}}$ and numerical attributes $\mathcal{J}_{\mathrm{cont}}$, according to their measurement scale.

\emph{Persona}: Given a population agent $x_i$, a persona $\pi_i$ is a textual description of its characteristics, automatically generated using an LLM based on the information provided by $x_i$. We denote the corresponding persona embedding as $\boldsymbol{e}_i$, obtained by encoding $\pi_i$ using a pretrained language model into a continuous semantic representation. The persona is not introduced merely as an alternative data representation, but as a semantic interface through which the generation process can be guided toward interpretable and policy-relevant target populations.

In our setting, the semantic-conditioned population synthesis task is to learn a generator $G_\theta(z, \boldsymbol{e})$ that models an implicit joint distribution $\hat{P}(X)$, where $z$ denotes random noise typically sampled from a standard Gaussian distribution $\mathcal{N}(0, I)$. The objective of $G_\theta(z, \boldsymbol{e})$ is to approximate the true population-level joint distribution $P(X)$ from a limited individual-level survey dataset of size $n \ll N$, while allowing the synthesized agents to be steered by semantic control signals.

\subsection{From Unconditional to Semantic-Conditioned Population Synthesis}

\begin{figure}[ht]
\centerline{\includegraphics[width=0.8\textwidth]{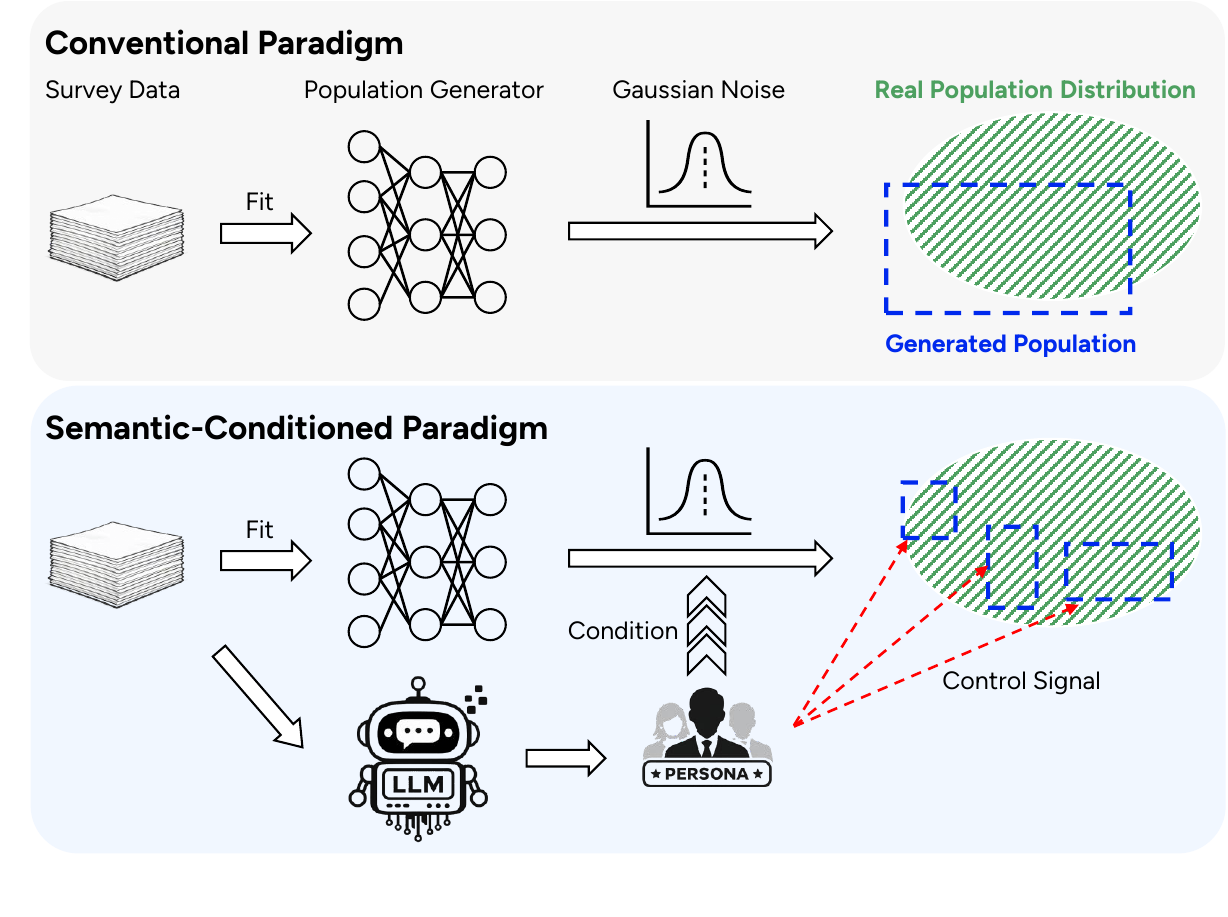}}
\caption{
Comparison between conventional generative population synthesis and our proposed semantic-conditioned paradigm. \emph{Top}: Conventional data-driven methods learn a generator from survey data and sample from Gaussian noise to approximate the overall population distribution, with limited controllability. \emph{Bottom}: We introduce a semantic-conditioned modeling paradigm, where persona text derived via an LLM serves as a control signal. This enables the generator to be guided toward specified target subpopulations in a controllable and interpretable manner.
}
\label{fig_prob_bg}
\end{figure}

As illustrated in Figure~\ref{fig_prob_bg}, existing deep generative approaches to population synthesis can be viewed as \emph{unconditional distribution learning frameworks}. Methods such as VAEs, GANs, and diffusion models learn a generator $p_{\theta}(x)$ from survey data to approximate the joint population distribution, and generate synthetic individuals by mapping latent noise $z$ to the data space:
\begin{equation}
x \sim p_{\theta}(x), \quad z \sim \mathcal{N}(0, I).
\end{equation}
This paradigm is effective when the goal is to reproduce the overall empirical distribution. However, it offers only limited control over \emph{which} parts of the distribution are sampled. In practice, this means that unconditional generation is well suited for average-case reconstruction, but less suitable for applications requiring targeted synthesis of rare, policy-sensitive, or counterfactual subpopulations.

For example, transport planning often requires synthetic populations reflecting specific planning assumptions, such as a larger share of transit-oriented commuters, greater prevalence of low-car-access households, or scenario-specific combinations of socio-demographic and behavioral traits. Such requirements are difficult to express through noise-only generation. As a result, existing methods remain primarily \emph{distribution-matching rather than distribution-controllable}, which limits their usefulness for scenario-driven population synthesis and policy analysis.

To address this limitation, we propose a \emph{semantic-conditioned population synthesis paradigm}, which extends the generative process to
\begin{equation}
x \sim p_{\theta}(x \mid \pi, z).
\end{equation}
Instead of relying solely on latent noise, our framework introduces persona-based conditioning, where structured population attributes are transformed into semantic persona text $\pi$ via an LLM and injected into the generator as control signals. This enables the model to steer generation toward specific regions of the population distribution, thereby supporting controllable synthesis of target subpopulations while retaining statistical realism.

A key design choice is to use \emph{persona text} as the conditioning signal, rather than raw attributes or clustered representations. By converting structured survey data into natural language descriptions, we obtain a semantically coherent representation that can unify heterogeneous information across demographic, household, and behavioral dimensions. Compared with attribute-level conditioning, persona text offers greater flexibility in expressing high-level assumptions, composite traveler profiles, and scenario narratives that may be difficult to encode through a fixed set of structured variables alone. This makes the conditioning mechanism more interpretable and more adaptable for target-oriented and scenario-driven population synthesis. In this sense, personas function as a semantic control layer between structured survey data and generative modeling, allowing the model to remain anchored in the learned data distribution while becoming more responsive to planning and policy needs.

Such controllable synthesis is particularly relevant for transport applications in which analysts need to evaluate populations under targeted assumptions rather than historical averages, for example when assessing equity-sensitive traveler groups, testing behaviorally shifted populations under new policies, or constructing rare but plausible subpopulations for simulation and forecasting.

\section{Methodology}
\label{method}

\subsection{Overview of SemaPop Framework}
\begin{figure}[ht]
\centerline{\includegraphics[width=0.6\textwidth]{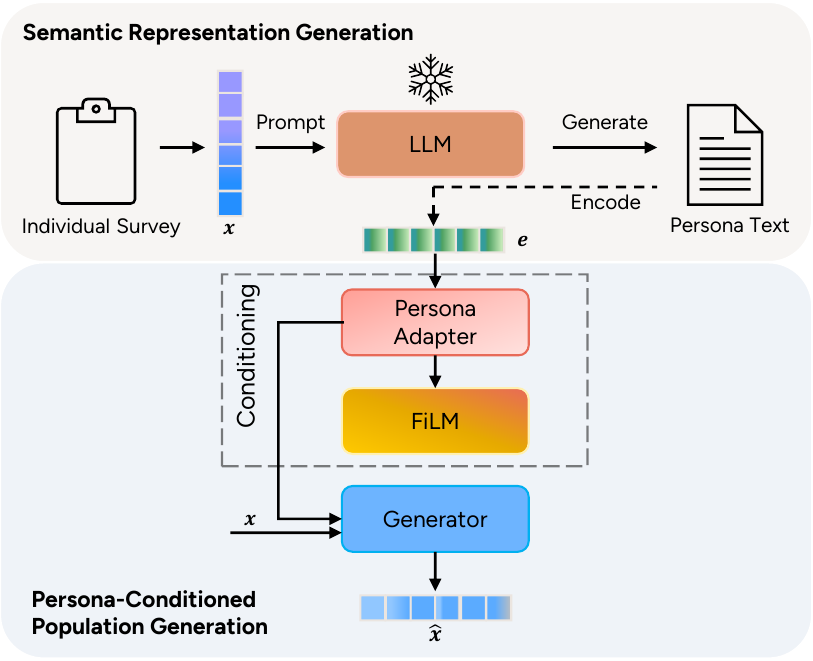}}
\caption{
Overview of the SemaPop framework. Individual survey data $x$ are used to prompt a frozen LLM to generate persona text, which is encoded into embeddings $\boldsymbol{e}$ and injected into the generator via a conditioning module (persona adapter with FiLM modulation). The generator learns to produce persona-conditioned agents $\hat{x}$ aligned with real data. During generation, persona embeddings combined with noise enable synthesis of new population agents.
}
\label{fig_frame}
\end{figure}

The framework shown in Figure \ref{fig_frame} operates in two phases, namely a training phase and a generation phase, each composed of modular components with distinct roles. The LLM functions as a semantic abstraction layer that transforms the structured information of population agents $x$ into high-level persona descriptions, while remaining frozen during training. The persona text is encoded by the same frozen LLM into a continuous semantic embedding, which serves as a compact representation of high-level persona information. The conditioning module injects this embedding into the generator, enabling persona-aware modulation of the generative process. During training, the generator learns to map persona-conditioned representations to realistic population agents under a prescribed training objective. During generation, the same modules are reused, with Gaussian noise replacing real data inputs, enabling the synthesis of new population agents guided solely by persona semantics.

\subsection{Semantic Representation Generation}

In population synthesis, individual agents are typically described by high-dimensional collections of demographic, household, and behavioral attributes. While such representations are suitable for statistical modeling, they are often difficult to interpret and insufficient to capture higher-level behavioral semantics. To address this limitation, we introduce personas as a semantic abstraction layer that summarizes heterogeneous attributes into interpretable, high-level descriptions of individual characteristics, following established practices in social science and transport research \citep{anable2005complacent, walker2002generalized, ben2002integration, cooper1999inmates, pruitt2003personas}.

In this study, personas are constructed by transforming structured survey attributes into natural language descriptions using a frozen large language model (LLM). The same LLM is then used to encode persona text into continuous embeddings, providing a compact semantic representation of individual-level characteristics. Formally, given a persona description $\pi_i$, we obtain a persona embedding
\begin{equation}
\boldsymbol{e}_i = \mathcal{E}_{\pi}(\pi_i),
\end{equation}
where $\mathcal{E}_{\pi}(\cdot)$ denotes a fixed embedding function instantiated by the frozen LLM.

To incorporate semantic information into the generative process while preserving the original tabular data space, we adopt feature-wise linear modulation (FiLM) \citep{perez2018film} as the conditioning mechanism. The persona embedding $\boldsymbol{e}$ is first transformed into a conditioning vector $\boldsymbol{c}$,
\begin{equation}
    \boldsymbol{c} = f_{\mathrm{adapt}}(\boldsymbol{e}),
\end{equation}
which is used to modulate intermediate feature representations in the generator via affine transformations:
\begin{equation}
    \boldsymbol{h}' = \bigl( \mathbf{1} + \boldsymbol{\gamma}(\boldsymbol{c}) \bigr) \odot \boldsymbol{h} + \boldsymbol{\beta}(\boldsymbol{c}).
\end{equation}

This design enables semantic information to influence generation in a continuous and differentiable manner without explicitly augmenting the underlying feature space. As a result, persona representations act as an independent semantic control signal, allowing the model to capture high-level behavioral patterns while maintaining consistency with structured statistical attributes.

\subsection{Marginal Regularization for Statistical Consistency}
\label{sec_margReg}
While the generative objective encourages realism at the individual level, it does not guarantee consistency with population-level statistics. In population synthesis, however, low-dimensional marginal distributions serve as primary validation targets and are critical for downstream scenario analysis. As a result, generative models that operate purely at the sample level may produce plausible individuals while exhibiting systematic deviations in aggregate distributions.

To address this limitation, we introduce marginal regularization as an explicit population-level constraint. Instead of enforcing sample-wise reconstruction, this regularization aligns generated samples with reference marginal distributions estimated from the training data. These reference marginals serve as internal statistical anchors that guide the model toward population-level consistency.

Formally, let $\hat X$ denote generated samples and $X_{\mathrm{ref}}$ denote reference data. For each variable $j \in \mathcal{J}$, we estimate the corresponding marginal distributions as
\begin{equation}
\hat p^{(j)} = \Phi_j(\hat X),
\qquad
p^{(j)} = \Phi_j(X_{\mathrm{ref}}),
\label{eq:hist_dist}
\end{equation}
where $\Phi_j(\cdot)$ denotes a marginal estimation operator.

We then define the marginal regularization term as
\begin{equation}
\mathcal{L}_{\mathrm{marg}}
= \frac{1}{|\mathcal{J}|}
\sum_{j \in \mathcal{J}} \mathrm{SRMSE}\!\left(\hat p^{(j)},\, p^{(j)}\right),
\end{equation}
where SRMSE measures the discrepancy between generated and reference marginals. The final training objective incorporates this term as a population-level constraint, encouraging the generator to remain statistically consistent while preserving individual-level realism.

\begin{figure}[htbp]
\centerline{\includegraphics[width=0.5\textwidth]{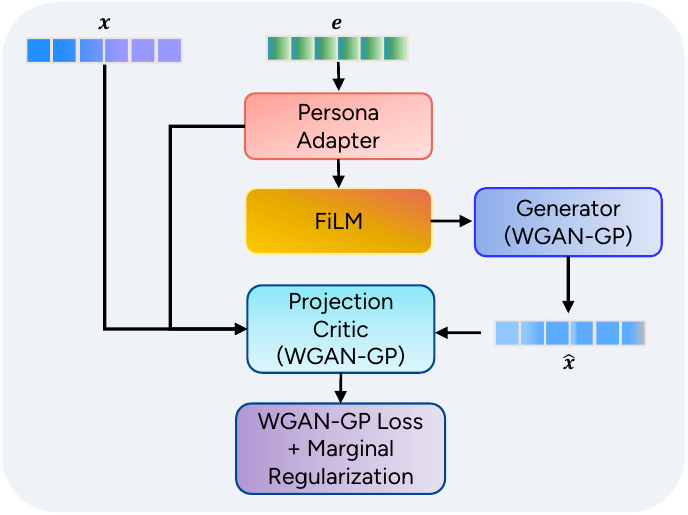}}
\caption{Overview of the SemaPop-GAN model. Persona embeddings condition a WGAN-GP based generator through FiLM, while marginal regularization enforces alignment with target population statistics under both factual and counterfactual scenarios.}
\label{fig_frame_gan}
\end{figure}

\subsection{Framework Instantiation: SemaPop-GAN}
\label{subsec:semapop_gan}

To instantiate the proposed framework, we develop \emph{SemaPop-GAN}, a persona-conditioned population synthesis model based on a conditional Wasserstein generative adversarial network (WGAN-GP) \citep{gulrajani2017improved}. This implementation serves as a concrete realization of the semantic-conditioned generation paradigm, rather than prescribing a specific generative architecture.

We adopt a GAN-based backbone due to its effectiveness in modeling high-dimensional tabular data and its flexibility in incorporating conditioning signals. In particular, the Wasserstein objective provides a stable training signal and aligns naturally with the goal of matching population-level distributions. As illustrated in Figure~\ref{fig_frame_gan}, persona embeddings are transformed via a lightweight adapter and injected into the generator through FiLM conditioning, enabling semantic information to modulate the generation process. To ensure consistency between generated samples and persona conditions, we further employ a projection discriminator \citep{miyato2018cgans}, which incorporates persona embeddings into the critic and enforces condition-dependent structural coherence. The generator is trained to jointly optimize adversarial learning and population-level statistical consistency. Specifically, its objective combines the adversarial loss with the marginal regularization term introduced in Section~\ref{sec_margReg}, encouraging the generated population to remain both semantically coherent and statistically aligned with reference distributions.

We emphasize that the SemaPop framework is model-agnostic. While SemaPop-GAN provides a concrete instantiation for empirical evaluation, alternative generative paradigms such as VAEs or diffusion models can be adopted, provided that they support conditional generation and population-level regularization.

\subsection{Counterfactual Interventions on Persona Condition}

While standard evaluation metrics (e.g., marginal fidelity or sample realism) assess the quality of generated populations, they do not reveal whether semantic conditioning provides effective control over the generation process. Under the proposed semantic-conditioned population synthesis paradigm, controllability is a defining property of the model. To explicitly diagnose this capability, we introduce counterfactual interventions on persona conditions, which probe whether changes in semantic inputs lead to systematic and interpretable shifts in the generated population. The key idea is to manipulate persona-related semantic inputs while holding other factors fixed, and to examine the resulting changes in generated populations. This allows us to isolate whether semantic information functions as an effective control signal rather than passively coexisting with statistical regularities.

\begin{figure}[htbp]
\centerline{\includegraphics[width=0.5\textwidth]{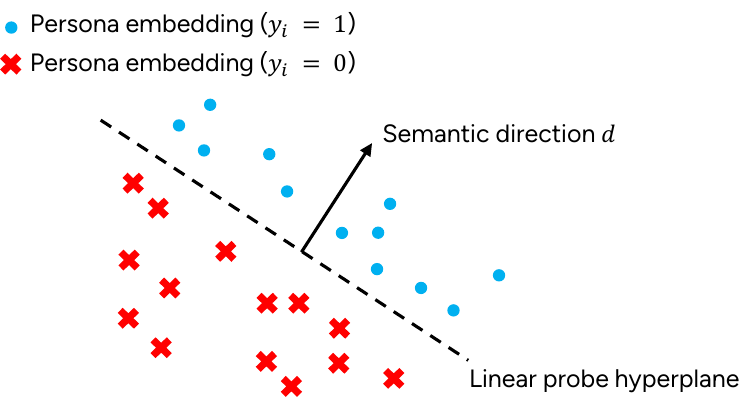}}
\caption{Semantic-level intervention in the persona embedding space. Persona embeddings are perturbed along a semantic direction while keeping latent noise fixed, enabling isolation of semantic effects on generated population agents.}
\label{fig_semaedit}
\end{figure}

\paragraph{Semantic-level intervention}
We perform counterfactual interventions directly in the persona embedding space. Given a persona embedding $\boldsymbol{e}_i$ and latent noise $\boldsymbol{z}_i$, the generated agent is
\begin{equation}
\hat{x}_i = G_\theta(\boldsymbol{z}_i, \boldsymbol{e}_i).
\end{equation}
To isolate the effect of semantic manipulation, we adopt a \emph{same-$z$} protocol, where the noise vector $\boldsymbol{z}_i$ is held fixed while the persona embedding is perturbed. Specifically, counterfactual embeddings are constructed as
\begin{equation}
\boldsymbol{e}_i^{(\alpha)} = \boldsymbol{e}_i + \alpha \boldsymbol{d},
\end{equation}
where $\boldsymbol{d}$ denotes a semantic direction and $\alpha$ controls the intervention strength. The resulting counterfactual samples are
\begin{equation}
\label{eq:semantic_edit}
\hat{x}_i^{(\alpha)} = G_\theta(\boldsymbol{z}_i, \boldsymbol{e}_i^{(\alpha)}).
\end{equation}
An illustration of this semantic perturbation process is provided in Figure~\ref{fig_semaedit}, where persona embeddings are edited along a semantic direction while keeping the latent noise fixed.

\begin{figure}[htbp]
\centerline{\includegraphics[width=0.9\textwidth]{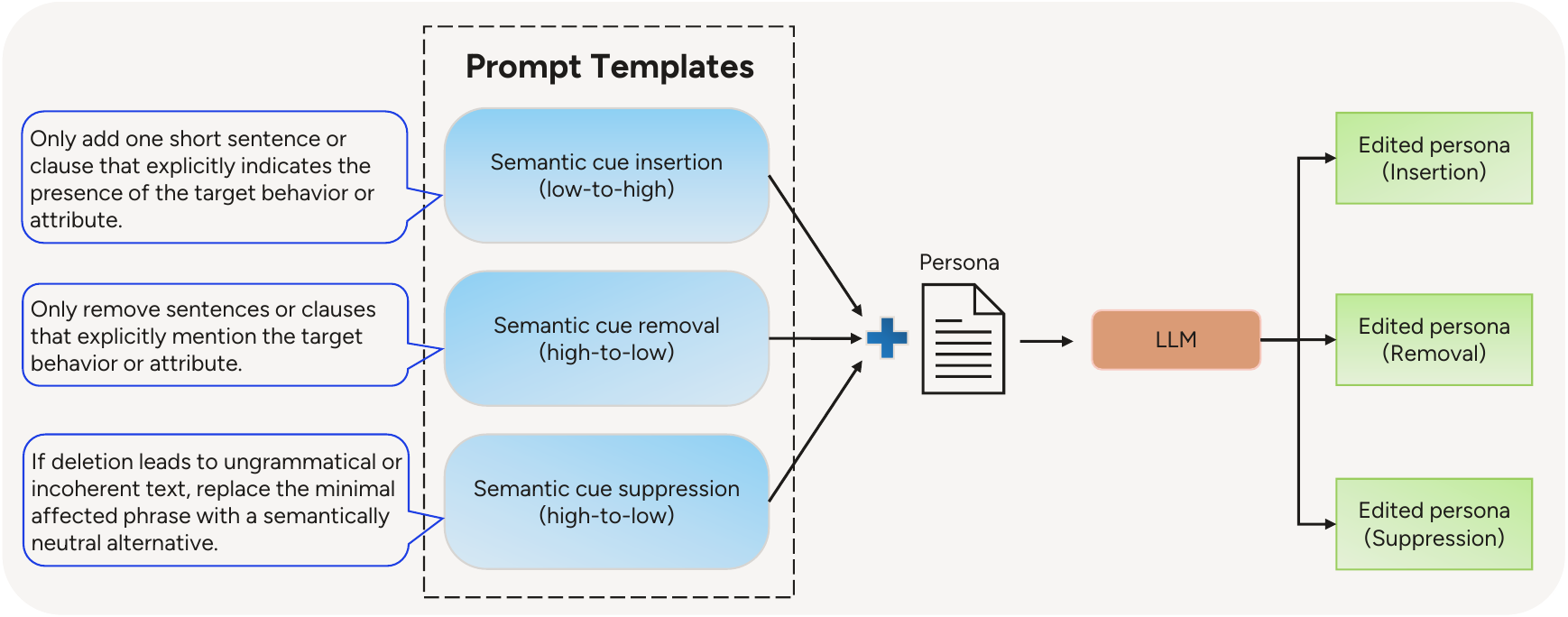}}
\caption{Text-level intervention via semantic editing of persona descriptions. Edited persona texts are re-encoded into embeddings and used to generate counterfactual population agents, enabling high-level semantic control.}
\label{fig_textedit}
\end{figure}

\paragraph{Text-level intervention}
We further consider text-level interventions by minimally editing persona descriptions and re-encoding them into embeddings. This provides a complementary high-level semantic interface for controlling generation without modifying the model architecture. Figure~\ref{fig_textedit} presents this intervention pipeline, where edited persona descriptions are re-encoded and used to generate counterfactual agents under the same latent noise.

Together, these interventions enable us to evaluate whether semantic conditioning induces systematic and interpretable shifts in the generated population distribution, which is essential for scenario-driven population analysis.

\section{Experiments}
\label{sec_ex}
\subsection{Data Description}
The experiments in this study are conducted using a large-scale synthetic population dataset representing the Swedish population, publicly available via Mendeley Data.\footnote{\url{https://data.mendeley.com/datasets/9n29p7rmn5/2}} 
The dataset provides a synthetic replica of over 10 million individuals (agents) in Sweden and is designed to support activity-based transport and population-level simulation studies. 
The synthetic population is generated on the basis of aggregated Swedish population and travel statistics for the year 2018, ensuring consistency with observed macro-level demographic and mobility patterns \citep{tozluouglu2023synthetic}. The data are organized in a relational format consisting of three linked tables: \emph{Person}, \emph{Household}, and \emph{Activity-travel}, capturing individual attributes, household context, and daily activity-travel patterns, respectively. 
The specific population attributes used as model inputs, together with their semantic grouping, data types, and numbers of distinct values, are summarized in Table~\ref{tab_popattrs}.

To construct the experimental population used in this study, we perform a stratified subsampling procedure at the municipality level. 
Sweden is divided into 290 municipalities, and from each municipality a fixed proportion of agents (e.g., 20\%) is randomly sampled from the synthetic population. This proportional sampling strategy preserves relative population sizes and spatial heterogeneity across municipalities while substantially reducing the computational burden of model training and evaluation. The resulting subsampled population is further partitioned into training, validation, and test sets with sizes of 50{,}873, 20{,}262, and 2{,}040{,}650 agents, respectively. This highly imbalanced split reflects a realistic population synthesis setting, in which population distributions must be inferred from a relatively small amount of available survey microdata and subsequently generalized to a much larger target population.

All experiments are conducted on the resulting subsampled population. Since the data are fully synthetic, they do not contain personally identifiable information and are used solely as a surrogate individual-level microdata source for methodological evaluation. Consistency with real-world population statistics is assessed through aggregate-level comparisons rather than direct individual-level validation.

\subsection{Evaluation Metrics}
\label{subsec:eval_metrics}
The evaluation metrics employed in this study serve two complementary purposes. First, they are used to assess the overall performance of the proposed generative model in reproducing statistically consistent synthetic populations. Second, they provide diagnostic measures for post-hoc marginal calibration, while counterfactual analyses rely on descriptive statistics of the relevant variables rather than additional formally defined evaluation metrics.

\vspace{0.5em}
\noindent
\textbf{Standardized Root Mean Square Error (SRMSE).}
In addition to its role in marginal regularization during model training, SRMSE is
formally defined here as an evaluation metric to quantify the similarity between the
generated and reference populations at the marginal and joint-marginal levels.
Given the empirical distributions
$\hat p^{(j)}=\Phi_j(\hat X)$ and $p^{(j)}=\Phi_j(X_{\mathrm{ref}})$ over the $j$-th
variable or variable subset, constructed using the aggregation operator
$\Phi_j(\cdot)$ defined in Eq.~\ref{eq:hist_dist}, SRMSE is defined as
\begin{equation}
\mathrm{SRMSE}\!\left(\hat p^{(j)},\,p^{(j)}\right)
=\frac{
\sqrt{\frac{1}{N_j}\sum_{k=1}^{N_j}\left(\hat p_{k}^{(j)}-p_{k}^{(j)}\right)^2}
}{
\frac{1}{N_j}\sum_{k=1}^{N_j} p_{k}^{(j)}
},
\end{equation}
where $N_j$ denotes the number of discrete categories or bin combinations for the
$j$-th marginal or joint-marginal distribution. 

In our experimental evaluation, we report \emph{SRMSE-M} when $j$ corresponds to a single variable, measuring consistency in univariate marginal distributions. We additionally report \emph{SRMSE-B} when $j$ denotes a pair of variables, which quantifies consistency in the corresponding bivariate joint distributions.

\noindent
\textbf{Feasibility and Diversity via Precision and Recall.}
Beyond distributional similarity, we further evaluate the feasibility and diversity
of the generated population using precision and recall \citep{kim2023deep}.
These metrics assess the mutual support between the generated samples $\hat X$ and the reference population $X_{\mathrm{ref}}$, which is used as the statistical anchor for evaluation.

Specifically, precision measures the proportion of generated samples that correspond
to attribute combinations observed in the reference population:
\begin{equation}
\mathrm{Precision}
= \frac{1}{|\hat X|} \sum_{i=1}^{|\hat X|}
\mathbb{I}\!\left(\hat x_i \in X_{\mathrm{ref}}\right),
\end{equation}
while recall measures the extent to which the generated population covers the
reference population:
\begin{equation}
\mathrm{Recall}
= \frac{1}{|X_{\mathrm{ref}}|} \sum_{j=1}^{|X_{\mathrm{ref}}|}
\mathbb{I}\!\left(x_j \in \hat X\right),
\end{equation}
where $\mathbb{I}(\cdot)$ is an indicator function that equals $1$ if the condition is satisfied and $0$ otherwise.

In this formulation, precision and recall provide indirect but informative measures of structural and sampling zeros. Low precision indicates the presence of infeasible or structurally implausible attribute combinations in the generated population (structural zeros), whereas low recall reflects insufficient coverage of valid but rare combinations observed in
the reference data (sampling zeros). Since precision and recall exhibit an inherent trade-off, we additionally report the F1 score as a balanced summary measure:
\begin{equation}
\mathrm{F1}
= \frac{2 \cdot \mathrm{Precision} \cdot \mathrm{Recall}}
{\mathrm{Precision} + \mathrm{Recall}} .
\end{equation}

\noindent
\textbf{Effective Sample Size (ESS).}
In post-hoc marginal calibration, the generated population is adjusted by assigning
non-uniform weights to individual agents in order to satisfy external marginal
constraints.
While such reweighting improves distributional consistency, it may also reduce the
effective diversity of the population by concentrating mass on a small subset of
agents.
To quantify this effect, we report the effective sample size (ESS), a widely used
diagnostic originally introduced in the context of importance sampling and
sequential Monte Carlo methods \citep{kong1992note,liu1998sequential}, and later adopted as a diagnostic of weight concentration and effective information
loss in reweighted populations \citep{deville1992calibration,biemer2012weighting}.

Given a calibrated population of $N$ agents with normalized weights $\{w_i\}_{i=1}^{N}$ satisfying $\sum_{i=1}^{N} w_i = 1$, ESS is defined as
\begin{equation}
\mathrm{ESS}
= \frac{1}{\sum_{i=1}^{N} w_i^2}.
\end{equation}
A higher ESS indicates that the calibrated population retains a larger proportion of effectively contributing agents, whereas a lower ESS reflects stronger weight concentration and reduced diversity. In our experiments, ESS serves as a diagnostic measure to assess the trade-off between marginal accuracy and population diversity induced by post-hoc marginal calibration.

\subsection{Generative Performance Comparison}
To evaluate the generative fidelity and statistical realism of the proposed \emph{SemaPop} framework, we conduct a comprehensive comparison against a diverse set of representative baseline models for tabular and population data generation. The selected baselines cover a broad spectrum of modeling paradigms, including classical probabilistic graphical models, VAE-based approaches, a diffusion-based model, and GAN-based generators. These baselines encompass both population synthesis methods and general-purpose tabular data generation models. All models are trained and evaluated under an identical experimental protocol and assessed using the same evaluation metrics, ensuring a fair and controlled comparison of their ability to reproduce individual-level joint distributions and population-level marginal statistics. Specific implementation details of the SemaPop are provided in ~\ref{app_imp_semapop}.

The comparison includes the following representative baseline methods:
\begin{itemize}
\item \textbf{CTGAN} \citep{xu2019modeling}.
A conditional GAN designed for tabular data generation, employing mode-specific normalization and conditional sampling to handle mixed discrete--continuous variables and imbalanced categorical distributions.

\item \textbf{TabDDPM} \citep{kotelnikov2023tabddpm}.
A diffusion-based generative model that adapts denoising diffusion probabilistic models to tabular data, providing a likelihood-based baseline for modeling complex joint distributions.

\item \textbf{TVAE} \citep{xu2019modeling}. 
A variational autoencoder tailored for tabular data generation, modeling the data distribution via a latent-variable formulation with reconstruction and KL-divergence objectives.

\item \textbf{BN} \citep{sun2015bayesian}. 
A Bayesian network--based population synthesis model that represents the joint distribution of individual and household attributes through a directed acyclic graph of conditional dependencies, offering a transparent and interpretable probabilistic baseline.

\item \textbf{BN-Copula} \citep{jutras2024copula}. 
A copula-based population synthesis BN model that decouples marginal distributions from dependency structures, enabling flexible modeling of continuous-variable dependencies and transferable synthetic population generation under varying marginal conditions.

\item \textbf{WGAN-GP} \citep{gulrajani2017improved}. 
A Wasserstein GAN baseline with gradient penalty regularization, which improves training stability and enforces the Lipschitz constraint more effectively than weight clipping in vanilla WGAN formulations.

\item \textbf{WGAN-GP-ZCR} \citep{kim2023deep}. 
A WGAN-GP-based population synthesis model that introduces novel generator-side regularization terms to explicitly account for structural zeros and sampling zeros, thereby improving the feasibility and convergence of synthetic populations in high-dimensional microdata.

\item \textbf{SemaPop-VAE}. Another instantiation of the SemaPop framework adopts a prior-conditioned variational autoencoder as the generative backbone, drawing on prior work on semantic-conditioned and learned-prior VAEs \citep{bowman2016generating,connor2021variational}. Implementation details are provided in \ref{app_frame_vae}.

\end{itemize}

\begin{table}[]
\centering
\caption{Generative performance comparison between the SemaPop and baseline models.}
\label{tab_perfcomp}
\begin{tabular}{llllll}
\hline
Model       & SRMSE-M     & SRMASE-B   & Precision      & Recall         & F1             \\ \hline
CTGAN       & 0.0119          & 0.0678          & 3.83           & 70.56          & 7.26           \\
TabDDPM     & 0.0118          & 0.0679          & 18.71          & 91.54          & 31.07          \\
TVAE        & 0.0119          & 0.0672          & 6.74           & 79.32          & 12.43          \\
BN          & 0.0119          & 0.0679          & 4.57           & 79.60          & 8.64           \\
BN-Copula   & 0.0119          & 0.0679          & 4.66           & 79.26          & 8.81           \\
WGAN-GP        & 0.0111          & 0.0607          & 52.01    & 93.08    & 66.74    \\
WGAN-GP-ZCR    & 0.0109    & 0.0593    & 51.88          & 93.33           & 66.69          \\
SemaPop-VAE & 0.0114          & 0.0628          & 15.20          & 86.10          & 25.84          \\
SemaPop-GAN & \textbf{0.0104} & \textbf{0.0554} & \textbf{73.95} & \textbf{93.39} & \textbf{82.54} \\ \hline
\end{tabular}
\end{table}

Table~\ref{tab_perfcomp} summarizes the generative performance of all models in terms of marginal error, bivariate consistency, and feasibility--diversity metrics.
Across baselines, marginal and bivariate errors remain tightly clustered, with SRMSE-M ranging from $0.0104$ to $0.0119$ and SRMSE-B from $0.0554$ to $0.0679$.
This limited variation indicates that marginal-level agreement alone provides insufficient discrimination of generative quality at the joint-distribution level.

In contrast, precision and recall reveal pronounced differences in feasibility and coverage.
Classical probabilistic models (BN and BN-Copula) and non-adversarial neural baselines (TVAE and TabDDPM) exhibit very low precision, all below $19\%$ (e.g., $3.83\%$ for CTGAN and $4.57\%$ for BN), despite achieving recall values in the range of $70$--$92\%$.
This pattern suggests that these models cover a substantial portion of observed attribute combinations but generate a large number of infeasible or structurally implausible samples, resulting in consistently low F1 scores below $32\%$.
GAN-based baselines substantially improve this trade-off: both WGAN-GP and WGAN-GP-ZCR raise precision to above $50\%$ while maintaining high recall around $93\%$, yielding F1 scores of approximately $66\%$.
The close performance between WGAN-GP and WGAN-GP-ZCR further indicates that stability-oriented regularization alone offers limited additional benefit once adversarial learning is in place.

A direct comparison between SemaPop-VAE and SemaPop-GAN highlights the importance of the generative backbone in translating semantic conditioning into feasible and diverse samples.
Although both models incorporate persona-aware conditioning and marginal regularization, SemaPop-GAN achieves substantially better feasibility and diversity, indicating a more faithful approximation of the joint population distribution at the sample level.
This gap can be attributed to fundamental differences in training objectives.
VAE-based models explicitly regularize the latent space toward a single Gaussian prior with a fixed parametric form, which often leads to over-smoothing of multi-modal dependencies in high-dimensional discrete attribute spaces. In contrast, WGAN-based models optimize a data-space distributional divergence without imposing a parametric latent prior. As a result, the WGAN-GP backbone provides a more expressive foundation for enforcing complex multivariate dependencies and suppressing structural zeros. This empirical contrast underscores the practical importance of backbone selection in population synthesis, rather than suggesting a universal superiority of adversarial models.

Overall, SemaPop-GAN achieves the best performance across all metrics, attaining the lowest SRMSE-M ($0.0104$) and SRMSE-B ($0.0554$) while simultaneously achieving the highest precision ($73.95\%$), recall ($93.39\%$), and F1 score ($82.54\%$).
Compared to the strongest GAN-based baselines (WGAN-GP and WGAN-GP-ZCR), SemaPop-GAN improves precision by over $22$ percentage points and F1 by more than $16$ points, while maintaining a comparable level of recall. This improvement indicates a substantial reduction in structurally implausible attribute combinations without compromising coverage of rare but valid configurations in the reference population.

\subsection{Ablation Studies on Semantic Inputs and Architectural Mechanisms}

To isolate the effect of semantic persona inputs, we ablate the form and semantic grounding of the persona provided to the generator by considering four persona settings, as summarized in Table~\ref{tab_ab_pers}. The results show clear and consistent differences across configurations. Removing semantic inputs (No Persona) leads to reduced structural validity, reflected by lower precision and F1, even though recall remains high, indicating that coverage alone does not prevent structurally implausible combinations. Using Randomized Persona texts further degrades precision and F1, suggesting that semantically fluent but individual-agnostic personas can increase structural inconsistencies when semantic alignment is absent. In contrast, both Attribute-Grounded Persona and Implicit Semantic Persona substantially improve structural consistency and coverage. While explicit personas achieve the highest precision and F1 at the cost of larger marginal errors, implicit semantic personas provide a more balanced trade-off between marginal accuracy and structural realism, as adopted in SemaPop.

\begin{table}[htbp]
\centering
\caption{Ablation results under four persona settings.
The four settings are: \textbf{No Persona} (zero persona embedding), \textbf{Randomized Persona} (LLM-generated personas without individual conditioning), \textbf{Attribute-Grounded Persona} (LLM-generated personas explicitly conditioned on precise individual-level attributes), \textbf{Implicit Semantic Persona} (free-form LLM-generated personas that intentionally abstract away precise attribute values, used in SemaPop).}
\label{tab_ab_pers}
\begin{tabular}{llllll}
\hline
Persona   Type                        & SRMSE-M & SRMSE-B & Precision & Recall & F1    \\ \hline
No Persona                            & 0.0106  & 0.0572   & 60.53     & 92.87  & 73.29 \\
Randomized Persona                    & 0.0112  & 0.0613   & 45.92     & 92.33  & 61.33 \\
Attribute-Grounded   Persona & 0.0137  & 0.0704   & \textbf{78.18}     & 93.03  & \textbf{84.96} \\
Implicit Semantic Persona             & \textbf{0.0104}  & \textbf{0.0554}   & 73.95     & \textbf{93.39}  & 82.54 \\ \hline
\end{tabular}
\end{table}

\subsection{Post-hoc Marginal Calibration}
\label{subsec:post_hoc_marg_cali}
To assess the robustness of persona-conditioned generation under explicit
distributional constraints, we conduct a post-hoc marginal calibration experiment.
The objective is to isolate whether population-level deviations induced by semantic
conditioning can be attributed to marginal mismatch, and to characterize the
trade-off between marginal fidelity, joint realism, and population diversity under
increasing constraint strength.

In this experiment, given a fully specified set of target marginal distributions, generated samples are kept fixed and reweighted at analysis time using a raking-based calibration procedure. Calibration strength is varied across five levels (L0--L4), corresponding to 0, 5, 10, 20, and 40 raking iterations, respectively. We evaluate two model variants—SemaPop trained with marginal regularization and an otherwise identical model trained without marginal regularization—under identical calibration levels. Full algorithmic and experimental details of the post-hoc calibration procedure are provided in ~\ref{app_marg_cali}.

All calibration experiments are conducted on synthetic population microdata.
Target marginals are derived from the same synthetic reference population or
constructed as hypothetical constraints for controlled analysis, rather than taken
from external real-world statistics.
Accordingly, post-hoc calibration is used strictly as an analytical tool, not as a
mechanism for fitting the model to ground-truth population marginals.

\begin{figure}[htbp]
\centerline{\includegraphics[width=1.0\textwidth]{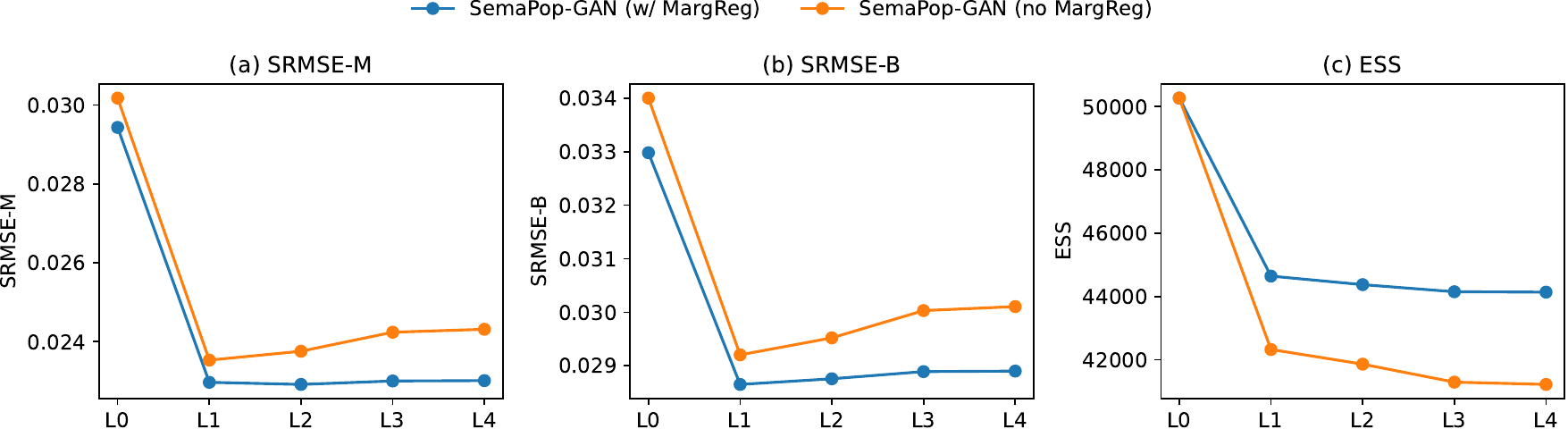}}
\caption{Trade-offs induced by post-hoc marginal calibration under increasing calibration strength (L0-L4). Panels report (a) marginal consistency (SRMSE-M), (b) joint realism (SRMSE-B), and (c) effective sample size (ESS). Results are shown for SemaPop-GAN trained with marginal regularization (\emph{w/ MargReg}) and an otherwise identical model trained without marginal regularization (\emph{no MargReg}).}
\label{fig_marg_cali}
\end{figure}

Figure~\ref{fig_marg_cali} reports marginal consistency (SRMSE-M), bivariate joint consistency (SRMSE-B), and effective sample size (ESS) across calibration levels.
Both models benefit from mild calibration (L1), but their behavior diverges as the calibration constraints strengthen. SemaPop-GAN with marginal regularization maintains stable marginal consistency and bivariate joint consistency from L1 to L4, indicating robust preservation of local joint-distribution structure under increasing constraints. In contrast, the unregularized model exhibits progressively degraded marginal consistency and bivariate joint consistency as calibration strength increases. Moreover, although ESS decreases for both models as expected, the regularized model consistently retains a higher effective sample size, indicating less severe weight concentration and improved robustness of the calibrated population.

These results suggest that marginal regularization during training not only improves
baseline marginal alignment, but also produces synthetic populations that better
accommodate subsequent distributional constraints, preserving both higher-order
structure and population diversity under semantic conditioning.

\subsection{Counterfactual Analysis on Semantic-Level Intervention}
To systematically evaluate the behavioral implications of semantic-level interventions, we conduct a series of counterfactual experiments using public transport (PT) usage as an illustrative intervened object. The choice of PT usage is not specific to the introduced counterfactual intervention framework, but serves as a representative behavioral attribute for examining how semantic manipulation propagates to observable outcomes. Specifically, this section investigates (i) population-level behavioral responses under continuous semantic perturbations, together with an analysis of potential side effects on non-target attributes, and (ii) heterogeneous intervention effects across distinct behavioral subgroups.

\paragraph{Evaluation set construction}
All counterfactual analyses in this study are conducted on a municipality-stratified evaluation set, referred to as the \emph{Municipality-Stratified Evaluation Set (MSES)}. The MSES is constructed by proportionally sampling individuals from the test set according to the population size of each municipality, resulting in a total of 10{,}054 agents. This construction follows the same municipality-level stratified sampling protocol adopted for the test set in the previous experiments, ensuring consistency in geographic composition and avoiding distributional discrepancies across evaluation settings.

\subsubsection{Continuous Semantic Intervention and Global Behavioral Response}
Semantic interventions are implemented by shifting persona embeddings along the learned intervention direction using a continuous scaling factor $\alpha \in \{-1.5, -1.0, -0.5, 0, 0.5, 1.0, 1.5\}$, as formalized in Eq.~\ref{eq:semantic_edit}. For each $\alpha$, public transport behavior is evaluated using two complementary metrics: the mean number of public transport trips, $\mathbb{E}\!\left[\texttt{Trips\_of\_PublicTransport}\right]$, capturing usage intensity, and the activation probability $P\!\left(\texttt{Trips\_of\_PublicTransport} > 0\right)$, capturing activation responses at the extensive margin. Side effects are further assessed by tracking variations in non-target attributes, enabling an evaluation of the selectivity and locality of the semantic intervention as the perturbation strength increases.

\begin{figure}[htbp]
\centering
\begin{minipage}[t]{0.46\textwidth}
    \centering
    \includegraphics[width=\linewidth]{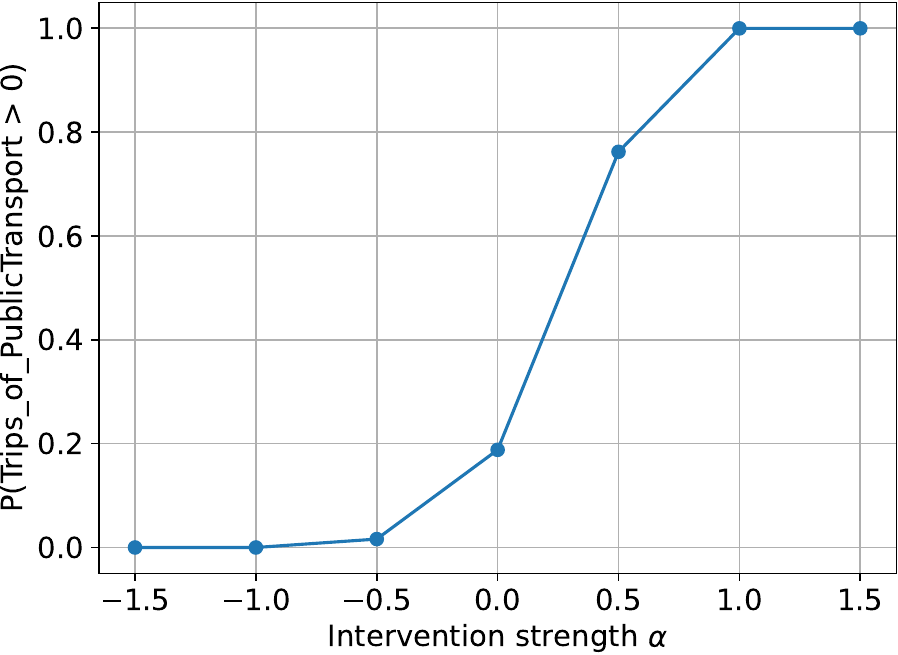}
    \par\vspace{2mm}
    \small (a) Activation (probability of at least one trip)

\end{minipage}
\hfill
\begin{minipage}[t]{0.46\textwidth}
    \centering
    \includegraphics[width=\linewidth]{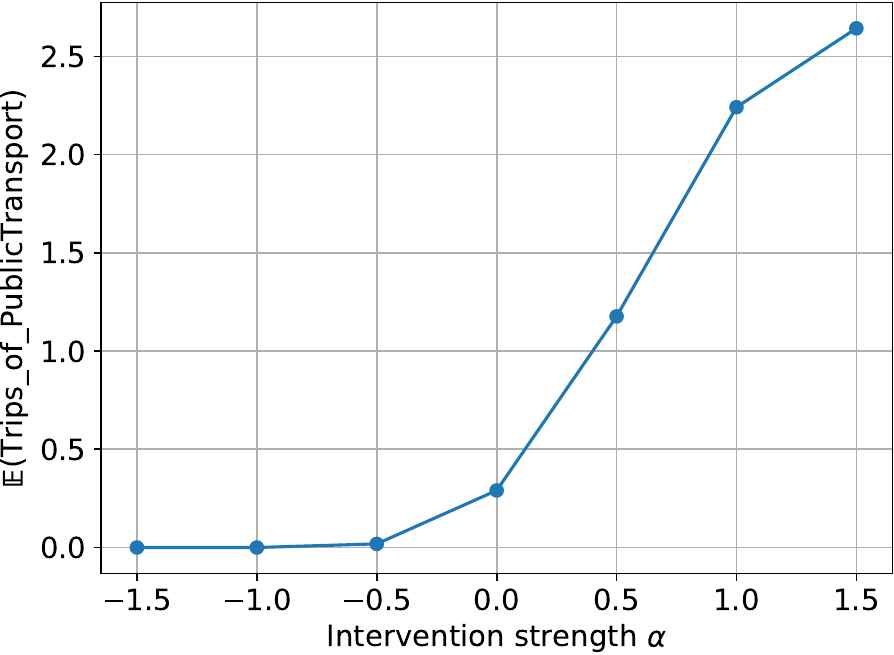}
    \par\vspace{2mm}
    \small (b) Intensity (expected number of trips)

\end{minipage}

\caption{Counterfactual response of public transport usage under semantic-level intervention.
(a) Probability of observing at least one public transport trip (activation).
(b) Expected number of public transport trips (intensity).
In both cases, the latent noise is held fixed to isolate the effect of semantic manipulation.}
\label{fig_count_pt}
\end{figure}

The results in Figures~\ref{fig_count_pt} reveal a coherent and monotonic behavioral response to semantic-level intervention. As the intervention strength $\alpha$ increases, both the activation probability of public transport usage and the expected number of public transport trips rise consistently, indicating that the semantic manipulation shifts agents toward higher public transport engagement. Figure~\ref{fig_count_pt}(a) highlights a strong extensive-margin effect: the probability of observing at least one public transport trip remains near zero for negative $\alpha$ values, but increases sharply beyond a positive threshold and quickly saturates. This pattern suggests that the intervention primarily activates latent public transport usage among previously inactive individuals. Conditional on activation, Figure~\ref{fig_count_pt}(b) shows a smooth increase in usage intensity with $\alpha$, indicating a clear intensive-margin response among active users. Throughout the intervention, the latent noise is held fixed, ensuring that the observed changes are driven by semantic manipulation rather than stochastic variation. Together, these results demonstrate that the learned semantic direction induces structured, interpretable, and population-level behavioral shifts, supporting its use for controlled counterfactual analysis.

\begin{figure}[htbp]
\centerline{\includegraphics[width=0.6\textwidth]{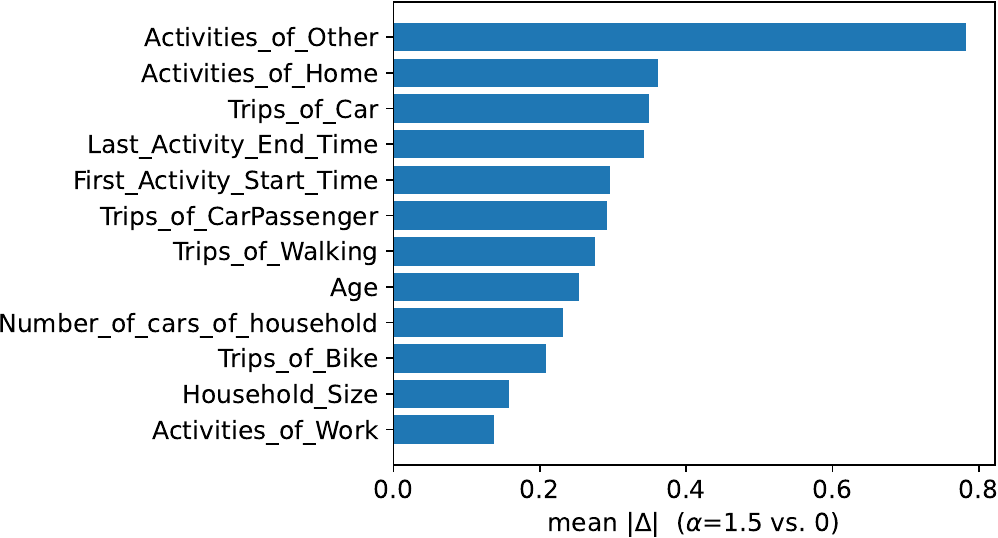}}
\caption{Locality and side-effect magnitude of semantic-level intervention.
The figure reports the mean absolute change of non-target numerical attributes between $\alpha=0$ and $\alpha_{\max}=1.5$, highlighting unintended perturbations induced by semantic manipulation while holding the latent noise fixed.}
\label{fig_count_se}
\end{figure}

Figure~\ref{fig_count_se} summarizes the side effects induced by semantic-level intervention by reporting the mean absolute changes of non-target numerical attributes. Overall, the magnitude of unintended perturbations remains limited and exhibits a clear locality pattern: attributes more semantically related to daily activity participation and mobility intensity (e.g., activity counts and car-related trips) show relatively larger responses, while core demographic and household attributes such as age and household size remain largely stable. This behavior indicates that the intervention primarily propagates within a semantically coherent subspace rather than inducing global, unstructured shifts across unrelated dimensions. Together with the monotonic target responses observed in Figure~\ref{fig_count_pt}, these results suggest that the semantic intervention achieves a favorable trade-off between behavioral effectiveness and side-effect control.

\subsubsection{Subgroup-Based Counterfactual Analysis}
\label{sec_count_subana}
While population-level analysis captures the average behavioral response to semantic intervention, it may obscure heterogeneous effects across individuals with different baseline behaviors. To examine whether semantic manipulation affects frequent and infrequent PT users differently, we further conduct a subgroup-based counterfactual analysis.

Subgroups are constructed from the MSES dataset based on observed PT usage. Specifically, the high-PT subgroup is defined as the top 10\% of individuals ranked by the number of public transport trips among those with strictly positive PT usage, which avoids instability caused by ties at zero. To form a contrasting low-PT subgroup of equal size, we adopt a mixed sampling strategy that combines individuals with low but positive PT usage and those with zero PT trips. Concretely, the low-PT subgroup consists of 5\% randomly sampled individuals from the PT-positive population and 5\% randomly sampled individuals with zero PT usage. The resulting high and low subgroups are non-overlapping and differ in size due to the underlying distribution of public transport usage, while remaining well separated in baseline behavior for comparative analysis. This procedure yields 1{,}005 high-PT and 770 low-PT individuals, which are used consistently across all subsequent counterfactual analyses, including the text-level interventions reported in the next subsection.

Table~\ref{tab_count_subana} illustrates heterogeneous counterfactual responses of PT usage across behavioral strata under semantic intervention. At the baseline ($\alpha=0$), the low-PT and high-PT subgroups are well separated in both activation and usage intensity, validating the subgroup construction. As the intervention strength increases, the two subgroups respond in opposite directions by design: positive semantic shifts activate and amplify PT usage among low-PT individuals, whereas reversed semantic shifts suppress and ultimately deactivate PT usage in the high-PT subgroup. The interplay between activation and intensity highlights the mechanism of intervention. For low-PT individuals, increased participation accompanies rising trip intensity, suggesting that semantic manipulation first activates latent PT usage and then reinforces usage levels. In contrast, the high-PT subgroup exhibits a strong extensive-margin contraction, indicating that the intervention effectively reduces PT participation among initially frequent users. 

Note that the activation is evaluated on the generated outcomes and therefore need not equal one even for the high-PT subgroup defined on the observed baseline data. This reflects the stochastic decoding process rather than a failure of subgroup definition. Overall, these results demonstrate that semantic interventions operate in a subgroup-dependent manner, inducing asymmetric responses across behavioral strata rather than uniform shifts at the population average.

\begin{table}[htbp]
\centering
\caption{Subgroup-specific counterfactual PT usage responses under semantic intervention. The row $\alpha=0$ denotes the unperturbed baseline. Positive $\alpha$ increases PT propensity for the low-PT subgroup, whereas the effective intervention direction is reversed for the high-PT subgroup, reflecting opposite semantic shifts across behavioral strata.}
\label{tab_count_subana}
\begin{tabular}{ccccc}
\hline
\multicolumn{1}{c|}{\multirow{2}{*}{$\alpha$}} & \multicolumn{2}{c}{Low-PT ($+\alpha$)} & \multicolumn{2}{c}{High-PT ($-\alpha$)} \\ \cline{2-5} 
\multicolumn{1}{c|}{}                       & mean      & activation     & mean      & activation      \\ \hline
0                                           & 0.36      & 0.32           & 1.86      & 0.98            \\
0.5                                         & 1.26      & 0.81           & 0.14      & 0.13            \\
1                                           & 2.23      & 1              & 0         & 0               \\ \hline
\end{tabular}
\end{table}

\subsection{Counterfactual Analysis on Text-Level Intervention}

To complement the semantic-level analysis, we examine text-level interventions that directly modify persona descriptions to assess whether explicit textual edits induce systematic behavioral changes across subgroups. The evaluation follows the same subgroup construction and sampling protocol as in Section~\ref{sec_count_subana}, with text-level interventions applied to the identical high-PT and low-PT subgroups defined based on observed PT usage in the original data (i.e., the pre-intervention baseline).

Table~\ref{tab_count_text} reveals a clear asymmetry in the effectiveness of text-level interventions across behavioral strata. For the high-PT subgroup, both Removal and Suppression edits lead to substantial reductions in public transport usage, as evidenced by large negative shifts in mean trip counts and activation probabilities. The median individual-level change of $-2$ trips, together with consistently negative 25th percentile, indicates that these effects reflect systematic downward shifts across individuals rather than being driven by isolated outliers.

Importantly, Suppression induces consistently stronger reductions than Removal across both aggregate and distributional metrics. Beyond the lower tail, this difference is also evident at the upper tail of the distribution. While Removal leaves the 75th percentile unchanged, Suppression produces a negative shift at the 75th percentile. This indicates a more uniform suppressive effect that extends to the highest-frequency PT users within the high-PT subgroup. One possible explanation is that simple removal of PT-related phrases may leave contextual cues in the persona text. These remaining cues can still encode frequent PT use, allowing high-frequency behavior to be partially preserved through the surrounding semantic context. In contrast, Suppression modifies the broader contextual framing of the persona description, enabling intervention effects to propagate more effectively to individuals whose travel patterns are strongly embedded in the original narrative.

By comparison, the Insertion intervention applied to the low-PT subgroup yields negligible changes across all metrics. Both the mean and activation probability remain essentially unchanged, and the distribution of individual-level trip differences is tightly concentrated around zero. This stark contrast suggests that text-level insertion primarily adjusts the magnitude of existing behavioral tendencies rather than creating new ones from scratch. These findings jointly suggest that text-level interventions are highly effective at suppressing entrenched behaviors among frequent users. However, they have limited capacity to activate PT usage among individuals with weak baseline travel behavior.

\begin{table}[htbp]
\caption{Counterfactual effects of text-level interventions on PT usage across behavioral subgroups. All $\Delta$ values are computed relative to the original personas, with P25/P75 denoting the 25th and 75th percentiles of individual-level trip differences.}
\label{tab_count_text}
\small
\begin{tabular}{lccccc}
\hline
Intervention                     & $\Delta$ Mean  & $\Delta$ Activation & \multicolumn{1}{l}{$\Delta$ Trips (Median)} & \multicolumn{1}{l}{$\Delta$ Trips (P25)} & \multicolumn{1}{l}{$\Delta$ Trips (P75)} \\ \hline
High-PT (Removal)     & -1.5443 & -0.4318      & -2                                   & -2                                & 0                                 \\
High-PT (Suppression) & -1.7473 & -0.4308      & -2                                   & -3                                & -1                                \\
Low-PT (Insertion) & -0.0234 & -0.0039      & 0                                    & 0                                 & 0                                 \\ \hline
\end{tabular}
\end{table}

\section{Conclusions}

This study introduces a \emph{semantic-conditioned population synthesis paradigm}, instantiated through a hybrid framework that integrates large language models (LLMs) with deep generative models. The proposed SemaPop framework derives persona representations from survey data and uses them as semantic conditioning signals, enabling controllable and interpretable generation of individual-level population agents under statistical constraints. We implement this paradigm using a WGAN-GP backbone (SemaPop-GAN) and incorporate marginal regularization to ensure alignment with target population distributions.

To address the practical constraints of large-scale real-world population surveys, all experiments are conducted using a synthetic Swedish population dataset. A diverse set of baseline models spanning multiple generative paradigms for population synthesis and tabular data generation is evaluated. The results consistently indicate that WGAN-GP-based models are better suited for population synthesis, offering stronger capacity to capture complex multivariate dependencies and suppress structural zeros. Building on this foundation, SemaPop-GAN further reduces marginal and joint distribution errors while improving sample-level feasibility and diversity. Ablation results highlight that implicit semantic personas achieve a more favorable balance between marginal consistency and structural realism compared to alternative persona designs.

In post-hoc marginal calibration, marginal regularization improves baseline marginal alignment while preserving higher-order structure and population diversity under semantic conditioning. Counterfactual analyses demonstrate that semantic-level interventions induce systematic and interpretable shifts in generated populations, with clear monotonic target responses. Subgroup analyses reveal asymmetric behavioral responses, where high-PT users react more strongly to suppression, while low-PT users exhibit weaker responses to incentive-based interventions. Furthermore, text-level interventions show that insertion minimally alters persona representations, whereas removal and suppression disrupt semantic structure, with suppression producing the strongest effects.

Overall, the results demonstrate that semantic conditioning enables controllable population synthesis beyond conventional unconditional generation. Beyond population synthesis, this paradigm motivates a shift toward generative modeling-based population projection, complementing conventional model-driven or assumption-based approaches. Future work will extend SemaPop to larger real-world population datasets and develop integrated population projection frameworks under dynamic demographic conditions, supporting scenario-driven population analysis and behaviorally informed forecasting in complex socio-economic systems.

\section{CRediT authorship contribution statement}
\textbf{Zhenlin Qin}: Conceptualization, Data curation, Methodology, Visualization, Investigation, Formal analysis, Validation, Writing – original draft.
\textbf{Yangcheng Ling}: Methodology, Validation,  Writing – review and editing.
\textbf{Leizhen Wang}: Data curation, Validation.
\textbf{Francisco Câmara Pereira}: Conceptualization, Formal analysis, Writing – review and editing, Supervision.
\textbf{Zhenliang Ma}: Conceptualization, Data curation, Methodology, Formal 
analysis, Writing – review and editing, Supervision.

\section*{Acknowledgements}
The work was supported by China Scholarship Council under Grant 202408320061, the TRENoP, and Digital Futures at KTH Royal Institute of Technology, Sweden.

\section*{Data availability}
The code and relevant resources for reproducing this study will be made publicly available at: https://github.com/qzl408011458/SemaPop.

\section*{Declaration of competing interest}
The authors declare that they have no known competing financial interests or personal relationships that could have appeared to influence the work reported in this paper.

\section*{Declaration of generative AI and AI-assisted technologies in the manuscript preparation process}
During the preparation of this work the author(s) used ChatGPT-5.2 in order to improve academic writing. After using this tool/service, the author(s) reviewed and edited the content as needed and take(s) full responsibility for the content of the published article.

\appendix
\section{Individual-Level Population Attributes}
\setcounter{table}{0}
\renewcommand{\thefigure}{\Alph{section}\arabic{table}} 

\begin{sidewaystable}[p]
\centering
\caption{Individual-level population attributes used as inputs to the generative model. Attributes are grouped by population semantics (demographic, household, and behavioral) and annotated with data type and number of distinct values.}
\label{tab_popattrs}
\small
\resizebox{1.0\linewidth}{!}{
\begin{tabular}{llllll}
\hline
\textbf{No.} & \textbf{Name}                   & \textbf{Attribute group} & \textbf{Data type} & \textbf{Number of values} & \textbf{Notes}                                                                   \\ \hline
1            & Municipality\_Categories        & Demographic              & Categorical        & 10                        & Residential municipality category, used as a spatial   demographic context       \\
2            & Age                             & Demographic              & Numerical          & 106                       & Individual age in years                                                          \\
3            & Gender                          & Demographic              & Categorical        & 2                         & Binary gender indicator                                                          \\
4            & Marital\_status                 & Demographic              & Categorical        & 3                         & Marital status of the individual                                                 \\
5            & Employment\_status              & Demographic              & Categorical        & 2                         & Employment participation status                                                  \\
6            & Studenthood\_status             & Demographic              & Categorical        & 2                         & Indicates whether the individual is a student                                    \\
7            & Income\_class                   & Demographic              & Categorical        & 5                         & Discretized personal income category                                             \\
8            & Household\_Type                 & Household                & Categorical        & 3                         & Household composition category                                                   \\
9            & Household\_Size                 & Household                & Numerical          & 61                        & Number of individuals in the household (discrete count)                          \\
10           & Number\_of\_cars\_of\_person    & Household                & Numerical          & 4                         & Household vehicle ownership attributed to individuals for   modeling convenience \\
11           & Number\_of\_children            & Household                & Numerical          & 39                        & Number of children in the household (discrete count)                             \\
12           & Number\_of\_cars\_of\_household & Household                & Numerical          & 26                        & Total number of vehicles owned by the household                                  \\
13           & Trips\_of\_Car                  & Behavioral               & Numerical          & 18                        & Daily number of trips conducted as car driver                                    \\
14           & Trips\_of\_CarPassenger         & Behavioral               & Numerical          & 17                        & Daily number of trips conducted as car passenger                                 \\
15           & Trips\_of\_PublicTransport      & Behavioral               & Numerical          & 17                        & Daily number of public transport trips                                           \\
16           & Trips\_of\_Walking              & Behavioral               & Numerical          & 18                        & Daily number of walking trips                                                    \\
17           & Trips\_of\_Bike                 & Behavioral               & Numerical          & 17                        & Daily number of bicycle trips                                                    \\
18           & Activities\_of\_Home            & Behavioral               & Numerical          & 7                         & Number of home activities in the daily schedule                                  \\
19           & Activities\_of\_Work            & Behavioral               & Numerical          & 7                         & Number of work-related activities                                                \\
20           & Activities\_of\_Other           & Behavioral               & Numerical          & 12                        & Number of non-work, non-home activities                                          \\
21           & Activities\_of\_School          & Behavioral               & Numerical          & 4                         & Number of school-related activities                                              \\
22           & First\_Activity\_Start\_Time    & Behavioral               & Numerical          & 1438                      & Start time of the first daily activity (minute resolution)                       \\
23           & Last\_Activity\_End\_Time       & Behavioral               & Numerical          & 1408                      & End time of the last daily activity (minute resolution)                          \\ \hline
\end{tabular}}
\end{sidewaystable}

This appendix documents the individual-level population attributes used as inputs to the generative modeling pipeline. Table~\ref{tab_popattrs} provides a comprehensive overview of all input attributes, including their population-semantic grouping, data type, and number of distinct values. All attributes are represented at the level of individual agents, while household-related characteristics are associated with individuals through their household membership and treated as shared contextual attributes. This classification is introduced for documentation and analytical transparency only and does not impose structural assumptions on the generative model.

For spatial context, the original 290 Swedish municipalities are grouped into 10 municipality categories, preserving essential spatial and urban--rural distinctions while reducing dimensionality. This grouping follows the official municipality classification scheme published by the Swedish Association of Local Authorities and Regions (SALAR), as documented in its municipality grouping report.\footnote{\url{https://skr.se/kommunerochregioner/kommungruppsindelning.8281.html}}

\section{Post-hoc Marginal Calibration}
\label{app_marg_cali}
\subsection{Scope and Role in the Analysis}
Post-hoc marginal calibration is employed as an analysis-time procedure to disentangle population-level effects arising from persona-conditioned generation from distributional shifts that may be attributed solely to marginal mismatch. The calibration procedure itself does not involve any persona intervention, but operates on fixed synthetic samples generated by the persona-conditioned model.

Specifically, post-hoc calibration adjusts the weights of generated samples to match a selected set of target marginal distributions, without modifying the underlying generative mechanism or semantic conditions. In the post-hoc marginal calibration experiment (Section~\ref{subsec:post_hoc_marg_cali}), the target marginal constraints are imposed on a selected set of key attributes capturing demographic structure (\textit{Age}), socioeconomic status (\textit{Income\_class}), household composition (\textit{Household\_Type}), motorization level (\textit{Number\_of\_cars\_of\_household}), and public transport usage intensity (\textit{Trips\_of\_PublicTransport}). These attributes are chosen to reflect dimensions that are commonly used in scenario-driven population analysis and policy-oriented studies. For example, public transport usage intensity may be indirectly influenced by household-level motorization, highlighting the importance of jointly considering demographic, household, and behavioral dimensions when assessing population-level responses under marginal constraints.

By isolating the influence of marginal constraints through reweighting, post-hoc calibration enables a controlled analysis of the trade-off between marginal fidelity (SRMSE-M), joint realism (SRMSE-B), and population diversity, as quantified by effective sample size (ESS). In this experiment, calibration strength is varied across five levels (L0--L4), corresponding to 0, 5, 10, 20, and 40 raking iterations, respectively.
\subsection{Calibration Problem Formulation}

Let $\{\hat{\mathbf{x}}_i\}_{i=1}^N$ denote a set of population agents generated by the
SemaPop model, where each agent is represented by a vector of categorical and numerical attributes.
For calibration, numerical attributes are discretized into bins, so that all variables can be treated uniformly as categorical.

The calibrated population is represented by a weighted empirical distribution
\begin{equation}
\hat{p}_{\mathbf{w}}(\mathbf{x})
= \sum_{i=1}^N w_i \, \delta(\mathbf{x} = \hat{\mathbf{x}}_i),
\end{equation}
where $\mathbf{w} = (w_1,\ldots,w_N)$ denotes non-negative sample weights.

Calibration is performed with respect to selected marginal constraints. For an attribute $j$ with categories $k \in \mathcal{K}_j$, let $p^{\ast}_j(k)$ denote the target marginal distribution. The corresponding weighted marginal induced by the synthetic population is
\begin{equation}
\hat{p}_{\mathbf{w},j}(k)
= \sum_{i=1}^N w_i \, \mathbb{I}\!\left(\hat{x}_{i}^{(j)} = k\right).
\end{equation}

The objective is to adjust $\mathbf{w}$ such that $\hat{p}_{\mathbf{w},j}(k) \approx p^{\ast}_j(k)$ for all constrained attributes, while keeping generated samples fixed.

\subsection{Raking-Based Calibration Procedure}

We employ an iterative proportional fitting (IPF) procedure to update sample weights. Starting from uniform initialization,
\begin{equation}
w_i^{(0)} = \frac{1}{N},
\end{equation}
weights are iteratively adjusted to reduce marginal discrepancies.

At iteration $t$, for attribute $j$, we compute a multiplicative adjustment ratio
\begin{equation}
r^{(t)}_j(k)
= \frac{p^{\ast}_j(k)}{\hat{p}_{\mathbf{w},j}^{(t)}(k) + \varepsilon},
\end{equation}
and update weights as
\begin{equation}
w_i^{(t+1)}
= w_i^{(t)} \cdot r^{(t)}_j\!\left(\hat{x}_{i}^{(j)}\right),
\end{equation}
followed by normalization to ensure $\sum_i w_i^{(t+1)} = 1$.

When multiple attributes are constrained, updates are applied sequentially within each iteration. This process is repeated for a fixed number of iterations, progressively aligning the generated population with target marginals.

\section{Framework Instantiation: SemaPop-VAE}
\label{app_frame_vae}
\setcounter{figure}{0}
\renewcommand{\thefigure}{\Alph{section}\arabic{figure}}

\begin{figure}[ht]
\centerline{\includegraphics[width=0.5\textwidth]{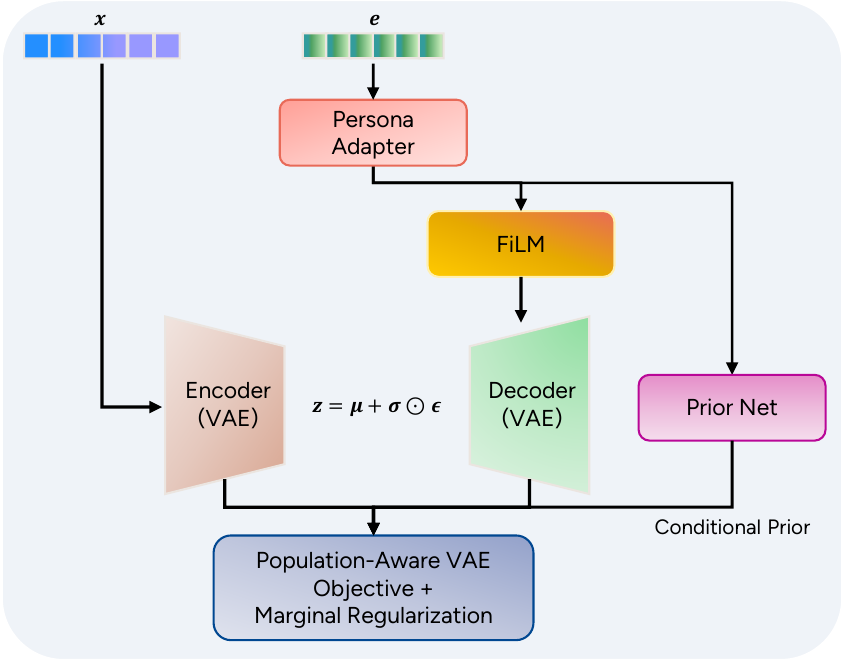}}
\caption{Overview of the SemaPop-VAE model. Persona embeddings condition the decoder through FiLM modulation and parameterize a conditional latent prior, while marginal regularization enforces population-level statistical consistency.}
\label{fig_frame_vae}
\end{figure}

As a reconstruction-based counterpart to SemaPop-GAN, we develop \emph{SemaPop-VAE}, a persona-conditioned variational autoencoder that provides an alternative instantiation of the proposed framework.

Persona embeddings are treated as external semantic signals and incorporated in two ways: (i) conditioning the decoder via FiLM to guide generation, and (ii) parameterizing a conditional latent prior to align the latent space with persona semantics. The encoder operates solely on observed attributes, ensuring that semantic information influences generation rather than inference.

The model is trained using a standard VAE objective augmented with marginal regularization:
\begin{equation}
\mathcal{L}
=
\mathcal{L}_{\text{rec}}
+
\beta\,\mathcal{L}_{\text{KL}}
+
\lambda_{m}\,\mathcal{L}_{\text{marg}},
\end{equation}
where $\mathcal{L}_{\text{rec}}$ denotes reconstruction loss for mixed-type attributes, $\mathcal{L}_{\text{KL}}$ is the KL divergence to a learned conditional prior, and $\mathcal{L}_{\text{marg}}$ is the marginal regularization term defined in the main text.

This formulation enables SemaPop-VAE to capture individual-level attribute distributions while enforcing population-level statistical consistency under semantic conditioning.

\paragraph{Discussion}
Compared to the adversarial instantiation, the VAE-based model tends to produce smoother distributions due to KL regularization, which may suppress rare but valid attribute combinations. As a result, while SemaPop-VAE provides a useful reconstruction-based baseline, it is less effective at capturing complex joint dependencies in high-dimensional population data. This observation highlights the importance of backbone choice for population synthesis under semantic conditioning.

\setcounter{figure}{0}
\renewcommand{\thefigure}{\Alph{section}\arabic{figure}}

\begin{figure}[htbp]
\centerline{\includegraphics[width=0.6\textwidth]{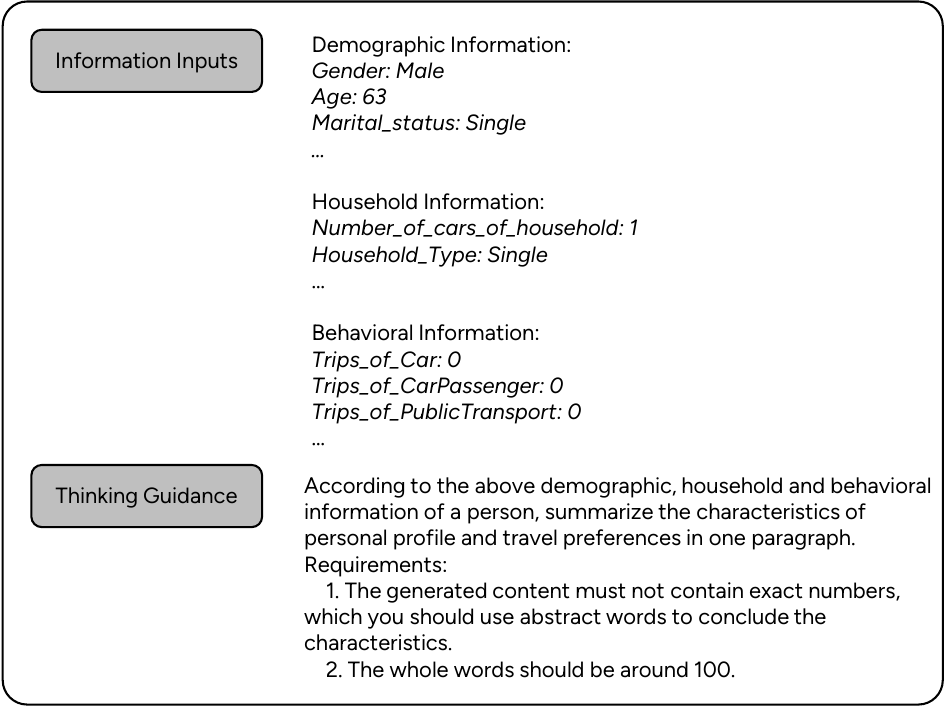}}
\caption{Prompt template for persona generation. The template specifies structured demographic, household, and behavioral inputs, together with explicit guidance and constraints, for prompting an LLM to generate persona descriptions. Content shown in brackets corresponds to an example for a specific population agent.}
\label{fig_promptpers}
\end{figure}

\section{Implementation Details of SemaPop}
\label{app_imp_semapop}
\begin{figure}[htbp]
\centerline{\includegraphics[width=0.4\textwidth]{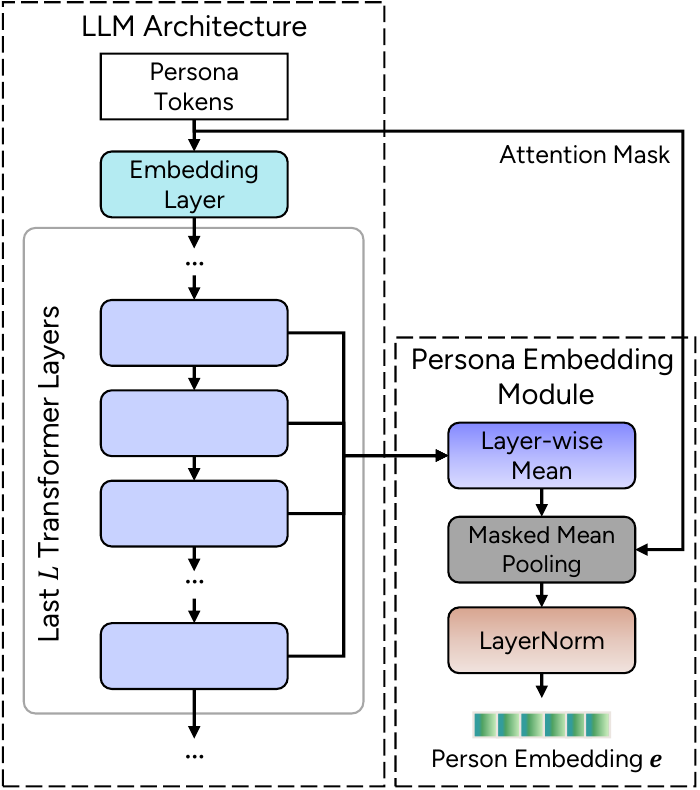}}
\caption{Persona Embedding Module. Persona text is first tokenized and processed by the LLM to obtain hidden states from the last $L$ Transformer layers. These layer-wise representations are averaged, followed by masked mean pooling over tokens and LayerNorm, to produce a fixed-dimensional persona embedding.}
\label{fig_embpers}
\end{figure}

To implement SemaPop, we employ a lightweight state-of-the-art large language model (LLM), Qwen3-8B, to generate persona descriptions from synthetic individual-level survey data with the prompt as shown in Figure~\ref{fig_promptpers}.
Persona generation is performed with a temperature of 0.9 and a top-$p$ value of 0.9, using a maximum of 512 newly generated tokens and with the thinking mode disabled. As illustrated in Figure~\ref{fig_embpers}, we extract high-level semantic information from the last \(L\) Transformer layers and aggregate them via a layer-wise mean operation. To ensure semantic consistency between persona generation and downstream conditioning, the same LLM is also used to obtain persona embeddings.

Persona generation constitutes the most time-consuming component of the pipeline, requiring approximately 4 seconds per persona. The persona embedding procedure is performed with a batch size of 4, resulting in a total processing time of approximately 3 hours to obtain embeddings for both the training and validation sets, comprising 71{,}135 samples in total. To improve overall pipeline efficiency, both persona generation and the persona embedding procedure are conducted as offline preprocessing steps, decoupled from the subsequent training of the SemaPop models. The main hyperparameters of the two SemaPop instantiations are reported in Tables~\ref{hyper_gan} and~\ref{hyper_vae}. 


All experiments are conducted on a workstation equipped with four NVIDIA RTX 6000 Ada GPUs, providing a total of 192~GB of GPU memory.

\setcounter{table}{0}
\renewcommand{\thetable}{\Alph{section}\arabic{table}} 

\begin{table}[htbp]
\centering
\small
\caption{Hyperparameters of SemaPop-GAN.}
\label{hyper_gan}
\begin{tabular}{ll}
\hline
Parameter               & Value           \\ \hline
Adapter hidden dim      & 1024            \\
Adapter condition dim   & 128             \\
Adapter dropout         & 0.1             \\
FiLM feature dim        & (256, 512, 256) \\
Generator hidden dim    & (256, 512, 256) \\
Critic hidden dim       & (256, 512, 256) \\
Gradient penalty weight & 10.0              \\
$\lambda_m$               & 0.4             \\
Generator learning rate & 2.0E-05        \\
Critic learning rate    & 2.0E-05        \\
Critic update frequency & 5               \\
Noise dim               & 128             \\
Batch size            & 512             \\
Epochs                & 300             \\
Optimizer               & Adam             \\ \hline
\end{tabular}
\end{table}

\begin{table}[htbp]
\centering
\small
\caption{Hyperparameters of SemaPop-VAE.}
\label{hyper_vae}
\begin{tabular}{ll}
\hline
Parameter             & Value           \\ \hline
Adapter hidden dim    & 1024            \\
Adapter condition dim & 128             \\
Adapter dropout       & 0.1             \\
FiLM feature dim      & (512, 512, 512) \\
PriorNet hidden dim   & 512             \\
Encoder hidden dim    & (512, 512, 512) \\
Decoder hidden dim    & (512, 512, 512) \\
Encoder dropout       & 0.1             \\
Decoder dropout       & 0.1             \\
$\beta$                  & 1.0               \\
$\lambda_m$             & 2.0               \\
Latent dim            & 128             \\
Learning rate         & 2.0E-04        \\
Batch size            & 512             \\
Epochs                & 300             \\
Optimizer             & Adam \\ \hline
\end{tabular}
\end{table}

\bibliographystyle{elsarticle-harv} 
\bibliography{_mybib}

\end{document}